\documentclass[sigconf]{acmart}

\usepackage{algorithm}
\usepackage{algorithmic}
\usepackage{amsmath}
\usepackage{multirow}
\usepackage{booktabs}
\usepackage{url}
\usepackage{graphicx}

\AtBeginDocument{%
  \providecommand\BibTeX{{%
    \normalfont B\kern-0.5em{\scshape i\kern-0.25em b}\kern-0.8em\TeX}}}

\copyrightyear{2022} 
\acmYear{2022} 
\setcopyright{rightsretained} 

\acmConference[SIGIR '22]{Proceedings of the 45th International ACM SIGIR Conference on Research and Development in Information Retrieval}{July 11--15, 2022}{Madrid, Spain.}
\acmBooktitle{Proceedings of the 45th Int'l ACM SIGIR Conference on Research and Development in Information Retrieval (SIGIR '22), July 11--15, 2022, Madrid, Spain}
\acmDOI{10.1145/3477495.3532080}
\acmISBN{978-1-4503-8732-3/22/07}

\usepackage{etoolbox}
\makeatletter
\patchcmd{\maketitle}{\@copyrightpermission}{
   \begin{minipage}{0.3\columnwidth}
     \href{http://creativecommons.org/licenses/by/4.0/}{\includegraphics[width=0.90\textwidth]{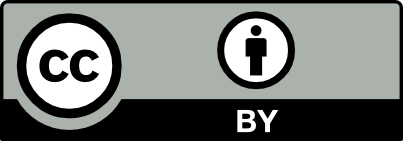}}
   \end{minipage}\hfill
   \begin{minipage}{0.7\columnwidth}
     \href{http://creativecommons.org/licenses/by/4.0/}{This work is licensed under a Creative Commons Attribution International 4.0 License.}
   \end{minipage}
  
   \vspace{5pt}
}{}{}

\makeatother

\begin{document}
\fancyhead{}
\title{What Makes the Story Forward? Inferring Commonsense Explanations as Prompts for Future Event Generation}
\author{Li Lin}
\affiliation{%
  \institution{Tsinghua University}
  \state{Beijing}
  \country{China}
}
\email{lin-l16@mails.tsinghua.edu.cn}

\author{Yixin Cao}
\authornote{Correspoding authors: Yixin Cao and Lijie Wen.}
\affiliation{%
  \institution{Singapore Management University}
  \country{Singapore}}
\email{caoyixin2011@gmail.com}

\author{Lifu Huang}
\affiliation{%
  \institution{Virginia Tech}
  \country{USA}}
\email{lifuh@vt.edu}

\author{Shu'Ang Li}
\affiliation{%
  \institution{Tsinghua University}
  \state{Beijing}
  \country{China}}
\email{lisa18@mails.tsinghua.edu.cn}

\author{Xuming Hu}
\affiliation{%
  \institution{Tsinghua University}
  \state{Beijing}
  \country{China}}
\email{hxm19@mails.tsinghua.edu.cn}

\author{Lijie Wen}
\authornotemark[1]
\affiliation{%
  \institution{Tsinghua University}
  \state{Beijing}
  \country{China}}
\email{wenlj@tsinghua.edu.cn}

\author{Jianmin Wang}
\affiliation{%
  \institution{Tsinghua University}
  \state{Beijing}
  \country{China}}
\email{jimwang@tsinghua.edu.cn}
\renewcommand{\shortauthors}{Li Lin, et al.}

\begin{abstract}
Prediction over event sequences is critical for many real-world applications in Information Retrieval and Natural Language Processing. Future Event Generation (FEG) is a challenging task in event sequence prediction because it requires not only fluent text generation but also commonsense reasoning to maintain the logical coherence of the entire event story. 
In this paper, we propose a novel explainable FEG framework, \textsc{Coep}. It highlights and integrates two types of event knowledge, sequential knowledge of direct event-event relations and inferential knowledge that reflects the intermediate character psychology between events, such as  {\it intents}, {\it causes}, {\it reactions}, which intrinsically pushes the story forward. To alleviate the knowledge forgetting issue, 
we design two modules, \textsc{Im} and \textsc{Gm}, for each type of knowledge, which are combined via prompt tuning.
First, \textsc{Im} focuses on understanding inferential knowledge to generate commonsense explanations and provide a soft prompt vector for \textsc{Gm}. We also design a contrastive discriminator for better generalization ability. Second, \textsc{Gm} generates future events by modeling direct sequential knowledge with the guidance of \textsc{Im}.
Automatic and human evaluation demonstrate that our approach can generate more coherent, specific, and logical future events.
\end{abstract}


\maketitle

\section{Introduction}
\begin{figure}
    \centering
    \includegraphics[width=\linewidth]{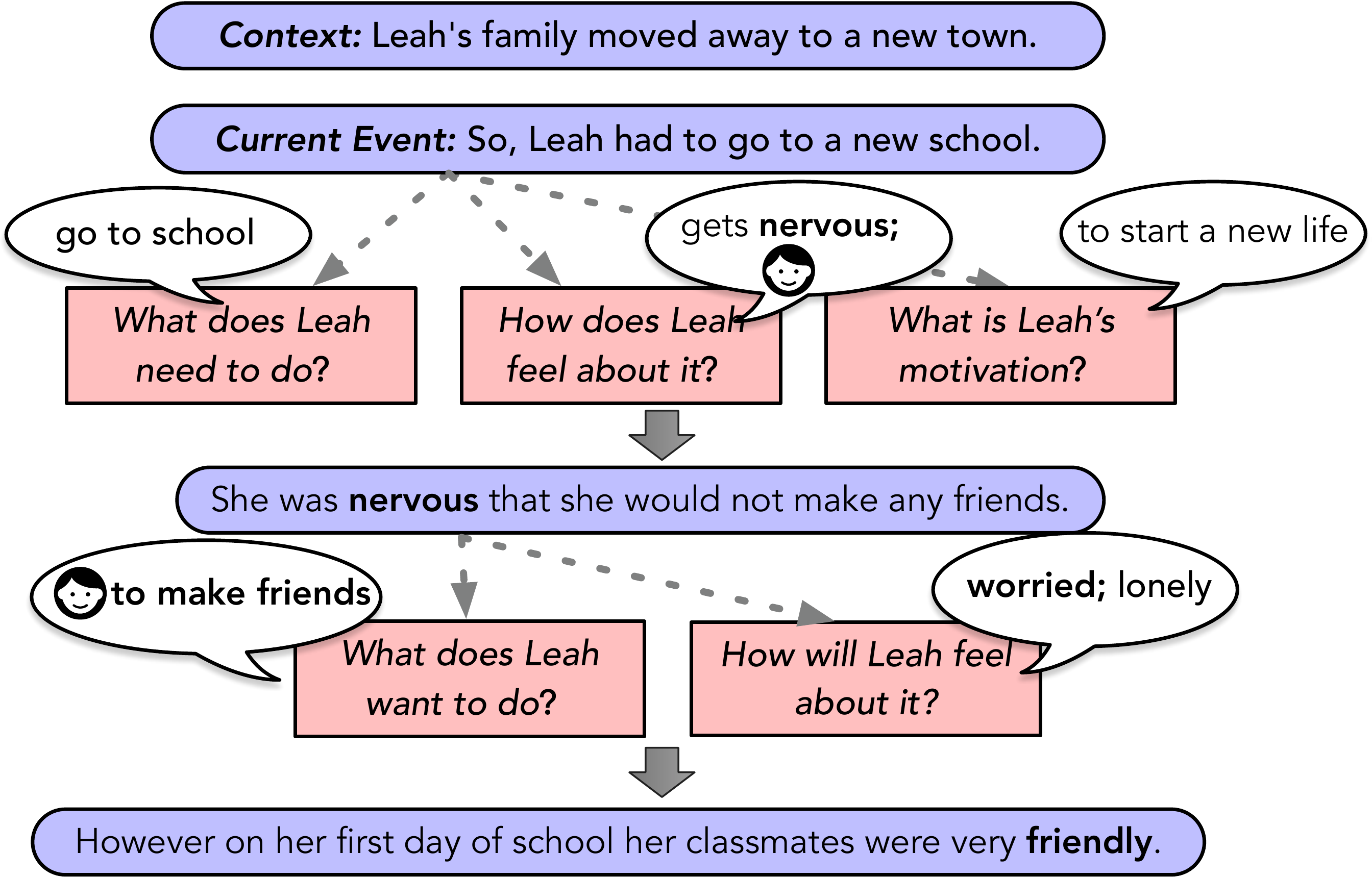}
    \caption{Examples of future event generation and commonsense explanation. The smiley faces indicate the dominant inferential knowledge for future events.}
    \label{fig:intro_example}
\end{figure}

Prediction over event sequences has many real-world applications in the fields of Information Retrieval and Natural Language Processing, such as Scholar Citation Forecasting~\cite{he2021cines}, Financial Quantitative Investments~\cite{cheng2020knowledge,feng2021hybrid}, Personalized Recommender Systems~\cite{kang2019deep}, Response Generation~\cite{Li2021KnowledgebasedRG,Li2021GraphStructuredCU}, and Story Telling~\cite{fan2018hierarchical,fan2019strategies,10.1145/3340531.3411937}. Focusing on Future Event Generation (FEG), it is a task of predicting the next event given preceding events, which requires to understand the inherent logical structure that pushes a story\footnote{In this work, a story is defined as a sequence of events, and each event is expressed by natural language.} forward. As exemplified in Figure~\ref{fig:intro_example}, given preceding events including the background context and current event, \textit{Leah moved to a new town} and \textit{she had to go to a new school}, a FEG system is expected to generate a consequence event, e.g. {\it she felt nervous about making new friends}.

With the help of pre-trained language models (PLMs), such as BERT~\cite{devlin2018bert},  GPT~\cite{radford2019language,brown2020language}, and BART~\cite{lewis-etal-2020-bart}, recent FEG studies are able to generate fluent sentences~\cite{goldfarb2020content}, but usually fail in capturing logical coherence among the events. The generated events are either too general --- ``he does not know'', or neutral with the preceding events ---producing ``go to the doctor'' given the preceding event ``to learn a sport''. How can we capture the endogenous relationship among sequential events to make it not only fluent but also reasonable?

In this paper, we investigate the logical states behind the story that pushes the event sequence forward. Inspired by ATOMIC~\cite{sap2019atomic}, we highlight the psychology of characters as a critical intermediate state between two adjacent events. That is, only if we understand the character's motivation, feeling, intention, etc., we can predict a coherent next event. As shown in Figure~\ref{fig:intro_example}, given that \textit{Leah had to go to a new school}, if we correctly infer that the \textit{emotional reaction} of \textit{Leah} would be \textit{nervous} and her \textit{motivation} is \textit{to start a new life}, we can better predict a future event, \textit{Leah felt nervous about making new friends}.
However, imposing logical constraints on the event generation process is challenging due to there are no available datasets providing sufficient psychology annotations. An efficient way to solve this issue is to leverage external commonsense Knowledge Graphs (KG).
A group of works~\cite{zhou2018commonsense,yang2019enhancing,ji2020language,xu2020megatron,10.1145/3340531.3417466} introduce abundant taxonomic (i.e. {\it AtLocation}, {\it IsA}, etc.) and event-event relations (i.e. {\it HasSubevent}, {\it HasPrerequisite}, etc.) from ConceptNet~\cite{speer2017conceptnet} as external information to enhance FEG models. The sequential knowledge expressed by the event-event relations can supplement the direct correlation between events as additional training data. Another line of works focus on the inferential knowledge in ATOMIC~\cite{sap2019atomic}. 
ATOMIC has organized with 9 typed {\it if-then} relations with variables (e.g., {\it ``if X repels Y's attack, X would intend to protect others from Y”}). These relations reflect an intermediate state --- indirect relations --- between events.
Clearly, the two types of knowledge are important and complementary to understand event sequence. However, few works utilize them both. We attribute one of the main reasons to the knowledge injection mechanism.
They fine-tune PLMs on both KGs and downstream tasks~\cite{guan2020knowledge} for knowledge transfer via shared parameters, while usually suffering from serious knowledge forgetting issues.

To this end, we propose a novel method that injects two types of knowledge for FEG, the inferential knowledge about the latent character psychology behind the event which can also provide explanations on how a story moves forward, and the sequential knowledge about the evolutional intrinsic among the events. We design two modules for each type of knowledge, \textsc{Im} and \textsc{Gm}, which are combined together via a prompting strategy. We name the framework as \textsc{Coep}\footnote{The code is available at \url{https://github.com/THU-BPM/CoEP}} that infers {\bf Co}mmonsense {\bf E}xplanations\footnote{We use {\it inferential knowledge} and {\it commonsense explanations} to express the latent psychology states which push an event story forward and the generated text scripts of the \textsc{Im} separately.} to {\bf P}rompt FEG. 
In specific, we initialize \textsc{Im} with BART~\cite{lewis-etal-2020-bart}, an state-of-the-art PLM specifically for text generation tasks. We then fine-tune it on ATOMIC~\cite{sap2019atomic} to capture inference knowledge similar to the prior studies~\cite{bosselut2019comet,hwang2021symbolic}. To alleviate the knowledge forgetting issue, the decoder will be frozen to generate intermediate explanations during FEG. The encoder outputs a soft prompt vector as the guidance for \textsc{Gm}. We also design a contrastive discriminator for \textsc{Im} to encourage the decoded explanation associate with input events for better generalization.
\textsc{Gm} targets generating the next event for FEG. We thus initialize it with a pretrained BART model and fine-tuned on the direct event-event knowledge --- the sequential KG derived from ConceptNet, which provides additional supervision.

In summary, the contributions of this work are:
\begin{itemize}
    \item We propose a novel \textsc{Coep} framework that integrates two types of commonsense knowledge. Except for direct sequential knowledge, \textsc{Coep} also highlights the reasoning of intermediate inferential knowledge between events and takes it to prompt a logically coherent future event.
	\item Our proposed \textsc{Coep} provides interpretability to FEG. Based on the generated explanations, we analyze the impacts of various types of character psychology on what makes the story forward to shed lights for future research (Section~\ref{sec:psychology anlysis}).
	\item We have conducted extensive experiments on publicly available benchmarks. Both automatic and human evaluations demonstrate the effectiveness of \textsc{Coep}, and further ablation studies on our results highlight the consistent, specific, and logical generation process.
 
\end{itemize}

\section{Related Work}

\subsection{Commonsense Inference}
Reasoning about the causes and effects of events with human commonsense is an important task in Natural Language Processing as it's an instinctive way to understand the law of logical events of humans~\cite{gordon2016commonsense}. Recently, as several benchmark datasets being constructed, such as {\it ROCStories} \cite{mostafazadeh-etal-2016-corpus, mostafazadeh2017lsdsem} which is a commonsense story corpus with partial mental state annotations and {\it wikiHow}, researchers have put more attention to understanding and generating the intents~\cite{rashkin-etal-2018-event2mind, rashkin-etal-2018-modeling}, and reasoning about relations between goals and steps ~\cite{zhang-etal-2020-intent, zhang2020reasoning}. In addition, several large-scale knowledge graphs are also constructed to support commonsense reasoning. For example, ConceptNet~\cite{speer2017conceptnet} is a knowledge graph that connects natural language words and phrases with labeled edges. It concentrates more on taxonomic knowledge instead of the knowledge of {\it how} and {\it why}. ATOMIC~\cite{sap2019atomic} focuses on inferential knowledge related to {\it if-then} reasoning. It defines nine {\it if-then} relations to distinguish causes vs. effects, agents vs. themes, and actions vs. mental states. Further, generative models, such as COMeT~\cite{bosselut2019comet,hwang2021symbolic}, are designed to automatically generate rich and diverse commonsense descriptions in natural language to make the knowledge graphs more complete. Inspired by these, we propose to train an automated inference model on ATOMIC which can generate commonsense explanations given the preceding events. It can naturally benefit the comprehension of commonsense reasoning of downstream FEG models, by taking the latent inferential knowledge as soft prompts to guide the generation of future events.

\subsection{Future Event Generation}
Pre-trained language models such as GPT~\cite{radford2019language,brown2020language}, BART~\cite{lewis-etal-2020-bart}, T5~\cite{raffel2019exploring} have shown the effectiveness in generation tasks such as text summarization~\cite{gupta2021automated} and machine translation~\cite{radford2019language}. Compared with such tasks of which the inputs have contained sufficient information to generate the desired output, future event generation is an open-ended generation task and especially requires commonsense inferences to generate logically consistent output. Previous studies on this task explored context clues from commonsense KG (e.g. ConceptNet) to introduce external understandings about the concepts and events~\cite{zhou2018commonsense,yang2019enhancing,ji2020language,xu2020megatron}. For example, given preceding events ``Mr. Egg was presenting a {\it volcanic} {\it eruption} to the science class.'', by reasoning on ConcepNet, it can be found that ``{\it eruption} is {\bf RelatedTo} {\it lava}'' and ``{\it volcano} is {\bf MadeOf} {\it lava}''. Then the generated ending event would be ``He shows the students that volcano exploded with substance that looked like {\it lava}!''~\cite{ji2020language}. Such efforts indeed help the language models generate more commonsense and informative texts in FEG, or story completion tasks, however, it is still a key challenge to keep the generated events logically coherent. Some works have observed that the language model would output counter-inferential events (e.g. given ``her favorite glasses were ruined'', the model would produce a causally deranged sentence: ``\dots {\bf then} she had a horrible accident. her sister had {\bf broken them}'')~\cite{xu2020megatron}. Thus, the inferential knowledge seems particularly important to keep the logical rhythm of story development, which explains the intention, motivation, or emotion of characters in the event.

Recent studies also proposed inferential KGs (ATOMIC) enhanced FEG models to generate reasonable and coherent stories~\cite{guan2019story,ammanabrolu2020automated,paul2021coins} by retrieving or generating inferential clues as additional input context but the coverage of inferential knowledge is lacking controlling. Some work introduced both ConceptNet and ATOMIC to fine-tune strong PLMs~\cite{guan2020knowledge} aiming to understanding both direct event-event information and indirect psychology knowledge to generate logical future events, but it is easily trapped into the knowledge forgetting issue when applied in downstream storytelling task. In stark contrast, our approach proposes to incorporate these two different KGs in a separate way, where one explicitly infers commonsense explanations about inferential knowledge behind events and the other one focuses on generating coherent future events based on the direct sequential knowledge and inferential understanding from the former one.

\subsection{Prompt Tuning}
Prompt tuning~\cite{brown2020language} is a simple yet effective mechanism for learning ``soft prompts'' from PLMs to perform specific downstream tasks. The prompts are usually continuous representations from a frozen model which typically refer to a task description and/or several canonical examples~\cite{shin2020autoprompt,reynolds2021prompt,li2021prefix,lester2021power}. There are two significant differences between our work and previous studies. First, instead of learning task-oriented prompts as previous studies did, we propose to generate all types of latent commonsense representations based on preceding events, where the total number of inferential relations are pre-defined in \textsc{Im} which is factually the 9 types of {\it if-then} relations in ATOMIC, and take them as instance-level prompts to guide FEG. Second, the prompts in our model are independent vectors attached to contextual representations of input events, while the prompts above are partial inner representations in pre-trained models (e.g., the prefix of hidden states in a layer). It can keep the commonsense prompts customized for each instance to make up for the problem of knowledge forgetting in fine-tuned models.

\section{Methodology}
\begin{figure*}
	\centering
	\includegraphics[width=\textwidth]{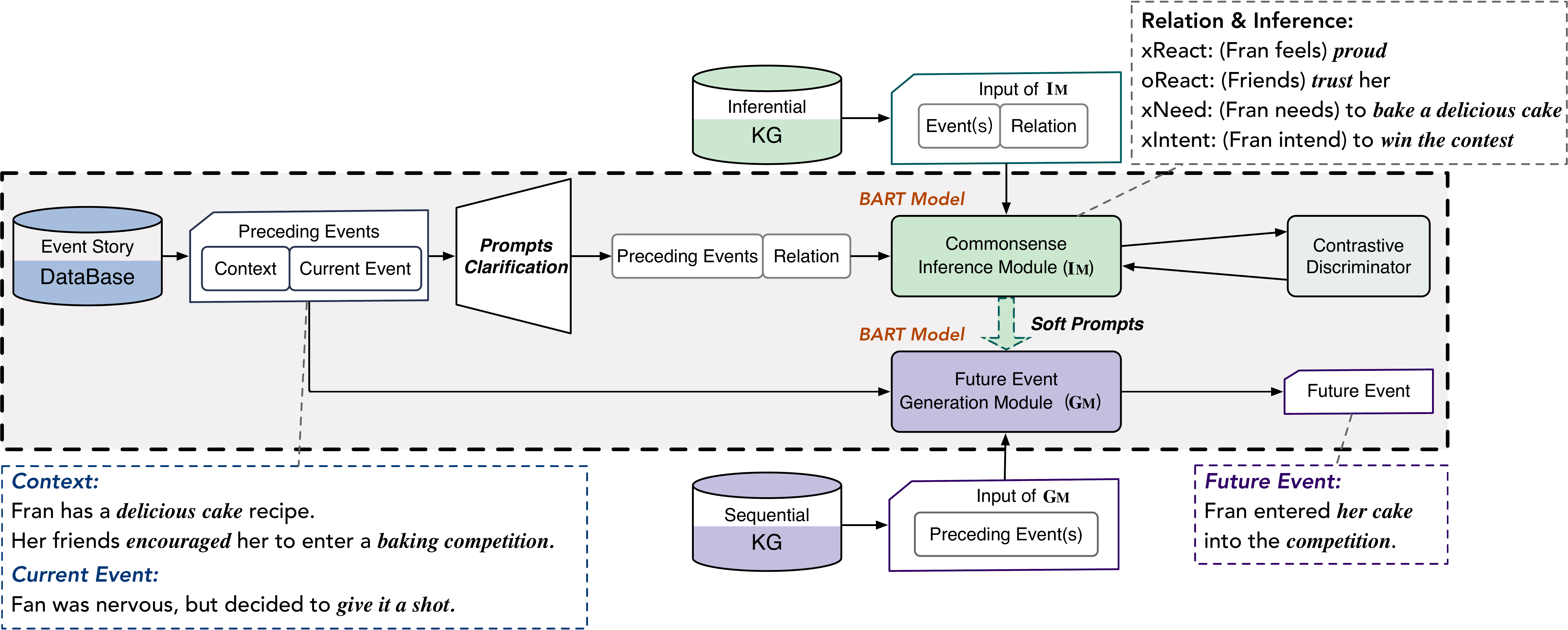}
 	\caption{The architecture of \textsc{Coep} framework. We decompose the framework into the following two parts: 1) the commonsense inference module (\textsc{Im}) fine-tuned on the inferential KG to provide commonsense reasoning; 2) the event generation model (\textsc{Gm}) to capture the contextual information of preceding events, which is fine-tuned on sequential KG. The inferential knowledge from \textsc{Im} is taken as soft prompts to guide future event generation in \textsc{Gm}, which is illustrated as the dashed arrow. A contrastive discriminator is created to distinguish whether an inferential explanation is consistent with the input event and the attached relation, which also provides weak supervision to constrain the coverage of inferential knowledge.
	}
	\label{arch}
\end{figure*}
We formulate the FEG task as follows: given a sequence of events\footnote{In event stories, each event is a sentence describing human's daily activities as shown in Figure~\ref{fig:intro_example}} $X=(e_1, e_2, e_3, \dots)$, in $t$-th generation step, $e_t$ indicates the current event, all history events $e_{1 \sim t-1}$ are considered as the background context and $e_{t+1}$ is the target future event the model should produce, the model learns to capture the contextual information and inferential knowledge from all preceding events and make a prediction of $e_{t+1}$. The generated event should be reasonable and logically consistent with the preceding events.

Our \textsc{Coep} framework aims to reason the inferential knowledge from preceding events to guide the FEG task using a prompting strategy. As shown in Figure~\ref{arch}, it consists of two components: 1) a commonsense Inference Module (\textsc{Im}), a pre-trained BART model fine-tuned on ATOMIC to infer the commonsense explanations of input events, which describe the mental states or causal factors of the characters in event scripts on a particular inferential relation (i.e., from 9 typed {\it if-then} relations as illustrated in Table \ref{tab:atomic}); and 2) a future event Generation Module (\textsc{Gm}), another BART model fine-tuned on a sequential KG that takes the various inferential knowledge as soft prompts to guide the FEG. Both of these two models are based on the large-scale pre-trained language model BART~\cite{lewis-etal-2020-bart} with the base settings.

Since BART model is a typical encoder-decoder form, we directly use the latent representations from \textsc{Im} encoder as continuous prompt vectors, which contains the inferential knowledge on various commonsense relations. 
Then the prompting vectors are injected to \textsc{Gm} for future event generation. To enforce the coverage of commonsense knowledge, we also design a contrastive discriminator to estimate the coherence between the commonsense explanations decoded from the prompting representations and the input preceding events. 

\begin{table}[h]
    \centering
    \begin{tabular}{lll}
        \toprule
        \multicolumn{3}{c}{Input Event: PersonX repels PersonY's attack} \\
        \midrule
        \textbf{xIntent} & \textbf{xEffect} & \textbf{oReact}  \\
        (PersonX intent) &  (PersonX effect) & (Other react) \\
        to protect others & gains an enemy & weak; ashamed \\
        \midrule
        \textbf{xNeed}  & \textbf{xWant}  & \textbf{oWant} \\
        (PersonX need) & (PersonX want) & (Other want) \\
        to defense himself & to call the police & attack again \\
        \midrule
        \textbf{xAttr}  & \textbf{xReact}  & \textbf{oEffect} \\
        (PersonX attribute) & (PersonX react) & (Other effect) \\
        skilled; brave & angry; tired & get hurts\\
        \bottomrule
    \end{tabular}
    \caption{An example of ATOMIC. Texts in () show the reformulated relations for \textsc{Im} fine-tuning in this paper.}
    \label{tab:atomic}
\end{table}

\subsection{Commonsense Inference Module}
As aforementioned, the commonsense Inference Module (\textsc{Im}) is based on a pre-trained BART~\cite{lewis-etal-2020-bart}. Following previous studies~\cite{bosselut2019comet, hwang2021symbolic}, we first fine-tune \textsc{Im} on ATOMIC~\cite{martin2018event}, a large-scale commonsense KG covering 9 types of inferential relations as described in Table~\ref{tab:atomic}. Such relations describe psychology states and causality between events, e.g. {\it Intent} indicates the character's intention, {\it React} reflects the emotion, and {\it Need} means the prerequiste. Given a triple from ATOMIC, $\langle e_{\text {head}}, r, e_{\text {tail}} \rangle$, where $e$ and $r$ represent a textual event and an inferential relation separately, we formulate the training pairs for \textsc{Im} as $\langle x_\mathcal{I}, y_{\mathcal{I}}\rangle$. $x_\mathcal{I}$ denotes a multi-segment sequence which concatenates an input event $e_{\text {head}}$ and the relational phrase $r$ corresponding to a particular inferential relation\footnote{We use the training splits from~\cite{sap2019atomic}, which splits 24,313 seed events into training, validation, and test sets (80\%/10\%/10\%), for fine-tuning \textsc{Im} where the average number of words in each event is 4.6.}, e.g., PersonX intent, as shown in the parenthesis in Table~\ref{tab:atomic}, while $y_{\mathcal{I}}$ indicates the target inferential explanation $e_{\text {tail}}$. For each relation in ATOMIC, we extend the term as ``[Agent role] [relation]'' to make it conform to natural language specification. For each segment in $x_{\mathcal{I}}$, we add two special tokens $\left\langle s \right\rangle$ and $\left\langle /s \right\rangle$ to represent the beginning and ending separately following~\cite{bhagavatula2019abductive}. We use $u$ to represent the textual output generated by the model decoders, {\text {ENC}} to indicate the encoder function, and {\text {DEC}} to indicate the decoder function in this paper. To fine-tune \textsc{Im} on ATOMIC, we autogressively predict the tokens in target scripts as shown in the following:
\begin{equation*}
    P(u_t|u_{<t}) = \sigma({\text {DEC}}_{\mathcal{I}}({\bf H}_{u_{<t}}^{d,l}, {\text {ENC}}_{\mathcal{I}}(x_{\mathcal{I}})){\bf W}+{\bf b})
\end{equation*}
where $u_t$ and $u_{<t}$ denote the $t$-th token and all the previous $t$-$1$ tokens in $u$. ${\bf H}_{u_{<t}}^{d,l}$ are the decoder hidden states of all the $t$-$1$ tokens from the $l$-th decoder layer. $l$ is the total number of layers in the encoder and decoder. $\text {ENC}_{\mathcal{I}}$ and $\text {DEC}_{\mathcal{I}}$ indicate the encoder and decoder in \textsc{Im} respectively. The matrix {\bf W} and {\bf b} are learnable parameters of linear function in the output layer, and $\sigma$ represents the softmax function to produce the probability of output tokens throughout this paper. The training objective is to minimize the following negative log-likelihood as the language modeling loss:
\begin{equation*}
    \mathit{L}_{\mathcal{I}}^{lm} = -\sum_{t=1}^{|u|}log P(u_t|u_{< t}) 
	\label{eq:lm loss}
\end{equation*}
in this loss function, $|u|$ denotes the total number of tokens in the target commonsense inference.

To better encourage the \textsc{Im} to infer the commonsense explanations, we further design a contrastive discriminator to score the coherence between the inference and the input event under a specific commonsense relation. For each training pair $s:\langle x_\mathcal{I}=(e_{\text {head}}, r), y_{\mathcal{I}}=e_{\text {tail}}\rangle$ constructed from ATOMIC, we randomly sample another $y'_{\mathcal{I}}$ from other tuples which have the same relation $r$ in the pairs $s': \langle x'_\mathcal{I}=(e'_{\text {head}}, r), y'_{\mathcal{I}}=e'_{\text {tail}} \rangle$ and construct a negative pair $\langle x_\mathcal{I}, y'_{\mathcal{I}}\rangle$. Thus, the training size for the contrastive discriminator is twice the raw training set. The discriminator is implemented based on the BART sequence classification head as a binary classifier to improve the contrastive comprehension of \textsc{Im}, which is further used to score the explanations decoded from the latent representations for soft prompts. The objective function to optimize the discriminator is in the following:

\begin{equation*}
\mathit{L}_{\mathcal{I}}^{D} = -log P(\mathbf{I}_s={\tilde{\mathbf{I}}_s}|s=\left \langle x, y\right \rangle)
\end{equation*}
\begin{equation*}
        \mathbf{I}_{s}= \left \{
            \begin{array}{ll}
            0, & s = \langle x_\mathcal{I}, y_{\mathcal{I}} \rangle\\
            1, & s = \langle x_\mathcal{I}, y'_{\mathcal{I}} \rangle
            \end{array}
        \right.
\end{equation*}
where $\mathbf{I}$ indicates the annotated label when constructing training samples and ${\tilde{\mathbf{I}}}$ refers to the binary logits produced by the discriminator. 

The overall objective of fine-tuning \textsc{Im} is to minimize the linear combination of the two objectives:
\begin{equation*}
\mathit{L}_{\mathcal{I}} = \mathit{L}_{\mathcal{I}}^{lm} + \mathit{L}_{\mathcal{I}}^{D}
\end{equation*}

\subsection{Event Generation Module}
The event Generation Module (\textsc{Gm}) aims to understand the contextual information from all preceding events combined with the inferential knowledge from \textsc{Im} to generate future events, which is also based on the pre-trained BART the same as \textsc{Im}. But the initial BART is pre-trained on open-domain corpus for language modeling tasks, there is still a gap between the raw BART model and our event generation model. To better acquire the FEG capability, we leverage the ConceptNet~\cite{speer2017conceptnet}, a general multilingual KG covering 36 relations, such as \textit{Antonym}, \textit{SimilarTo}, \textit{HasSubevent} and so on, where sequential relations are also included. We carefully select 6 types of relations related to sequential events as shown in Table~\ref{tab:conceptnet} and collect 39,530 event pairs $\langle e_p, e_f\rangle$ for fine-tuning \textsc{Gm}, where $e_p$ and $e_f$ denote the preceding and future event respectively. The triplets that contain counter-sequential relations would be reversed, i.e., the triplet $\langle \text{say ah ha, } {\it HasLastSubevent} \text{, create idea}\rangle$ will be formd as $\langle e_p\text{=create idea}, e_f\text{=say ah ha}\rangle$.
The average number of words in the events is 2.67. We still use $u$ to represent the output of the generation module. The objective of fine-tuning \textsc{Gm} on sequential events KG is to generate $e_f$ given $e_p$ by minimizing the following negative log-likelihood:
\begin{equation*}
    L^{cn}=-\sum^{|u|} \sigma(\text {DEC}_{\mathcal{G}}({\bf H}_{u_{<t}}^{d,l}, \text {ENC}_{\mathcal{G}}(e_p)){\bf W} + {\bf b})
\end{equation*}
where $|u|$ denotes the total tokens in target tail events $e_f$. $\text {ENC}_{\mathcal{G}}$ and $\text {DEC}_{\mathcal{G}}$ indicate \textsc{Gm} encoder and decoder respectively.

\begin{table}[h]
    \centering
    \begin{tabular}{lcl}
        \toprule
        Head Event & Relation & Tail Event \\
        \midrule
        riding bike & Causes $\gg$  & falling down \\
        get check up & CausesDesire $\gg$  & know if is healthy \\
        playing chess & HasSubevent $\gg$ & capture queen \\
        apply for job & HasFirstSubevent $\gg$ & fill out application \\
        get weapon & HasPrerequisite $\ll$ & advance into battle \\
        say ah ha & HasLastSubevent $\ll$ & create idea \\
        \bottomrule
    \end{tabular}
    \caption{Examples of sequential triples in ConceptNet. The arrows show the order direction of relations before we transform them into sequential format.}
    \label{tab:conceptnet}
\end{table}

After fine-tuning \textsc{Gm} on the sequential events, we finally train it on the FEG task by considering both the preceding events and the inferential knowledge from \textsc{Im}. To enrich the context information, \textsc{Gm} will organize the input sequence as a multi-segment sentence, including all the history events $e_{1\sim t-1}$ as the background context and the current event $e_t$.  By concatenating the above two parts of sequences, the input of \textsc{Gm} is represented with $x_{\mathcal{G}}$. The special tokens $\left\langle s \right\rangle$ and $\left\langle /s \right\rangle$ are also added at the beginning and ending of each segment. To incorporate the inferential knowledge from \textsc{Im}, we introduce a prompting strategy that collects the last hidden state of the final ending token $\langle /s\rangle$ from \textsc{Im} encoder based on each commonsense relation, which is usually used to represent the contextual understanding upon the entire input sentence~\cite{devlin2018bert}. Specifically, given the input sequence $x_{\mathcal{G}}$, we need to inject it into the \textsc{Im} to capture the inferential knowledge on the perspective of psychology and causality. Considering an inferential relation $r_i$ among the 9 typed relations in ATOMIC, we first construct the prompts clarification $x_{\mathcal{I}_i}$ as the input for \textsc{Im} by concatenating $x_{\mathcal{G}}$ and $r_i$. Then we obtain the prompting vectors by collecting the ending positional hidden state of the input from \textsc{Im} encoder as follows:
\begin{equation*}
h_{r_i} = \text {ENC}_{\mathcal{I}}(x_{\mathcal{I}_i})_{\left\langle /s \right\rangle}, i\in \left[1, 9\right]
\end{equation*}

We then take the 9 dimensional inferential prompts along with the context encoding of all preceding events from the \textsc{Gm} encoder as input to the \textsc{Gm} decoder and generate a future event:
\begin{equation*}
{\bf H} = [h_{r_1}, h_{r_2}, \dots, h_{r_9},\text{ ENC}_{\mathcal{G}}(x_{\mathcal{G}})] 
\end{equation*}
\begin{equation*}
    P(u_t|u_{<t}) = \sigma(\text{ DEC}_{\mathcal{G}}({\bf H}_{u_{<t}}^{d,l}, {\bf H}){\bf W}+{\bf b})
\end{equation*}
where $u_t$ is the $t$-th token in the target future event from the event story dataset. 

The objective of future event generation is to minimize the negative log-likelihood as follows:
\begin{equation*}
    \mathit{L}_{\mathcal{G}}^{lm} = -\sum^{|u|}log P(u_t|u_{< t}) 
\label{equation:lm loss}
\end{equation*}
However, there is still a problem that how could we maintain the coverage of inferential prompts for input events. We will introduce a weak supervision approach based on the contrastive discriminator to this issue in the next subsection.

We add an auxiliary classification layer similar to the contrastive discriminator in \textsc{Im} to improve the distinguishing comprehension about the sequential events of \textsc{Gm}. Given a FEG training sample $\langle x_{\mathcal{G}}, y_{\mathcal{G}}\rangle$, the negative pair is constructed by replacing $y_{\mathcal{G}}$ with a randomly sample event $y'_{\mathcal{G}}$, where $y'_{\mathcal{G}} \neq y_{\mathcal{G}}$.
The classification task is designed to distinguish whether a future event is sequentially consistent with the preceding events, whose objective function is calculated as follows:
\begin{equation*}
 \mathit{L}_{\mathcal{G}}^{cls} = -log P(\mathbf{I}_s={\tilde{\mathbf{I}}_s}|s=\left \langle x, y\right \rangle)
\end{equation*}
\begin{equation*}
\mathbf{I}_{s}= \left \{
\begin{array}{ll}
0, & s = \langle x_\mathcal{G}, y_{\mathcal{G}} \rangle\\
1, & s = \langle x_\mathcal{G}, y'_{\mathcal{G}} \rangle
\end{array}
\right.
\end{equation*}
And the overall training loss for FEG is:
\begin{equation*}
    L_{\mathcal{G}} = L_{\mathcal{G}}^{lm} + L_{\mathcal{G}}^{cls}
\end{equation*}

\subsection{Prompt Training Strategy}
As we use the latent continuous representations as soft prompts to guide the generation of future events, the next question is: {\itshape {How to supervise the prompts collecting?}} It is challenging because there are no available datasets containing sufficient annotations of both future events and the latent inferential knowledge in-between the events. We propose to solve this problem by taking advantage of the contrastive discriminator trained along with the \textsc{Im}, which aims to measure the coherence of the commonsense explanations to the input event under an inferential relation.

\begin{figure}
	\centering
	\includegraphics[width=0.8\linewidth]{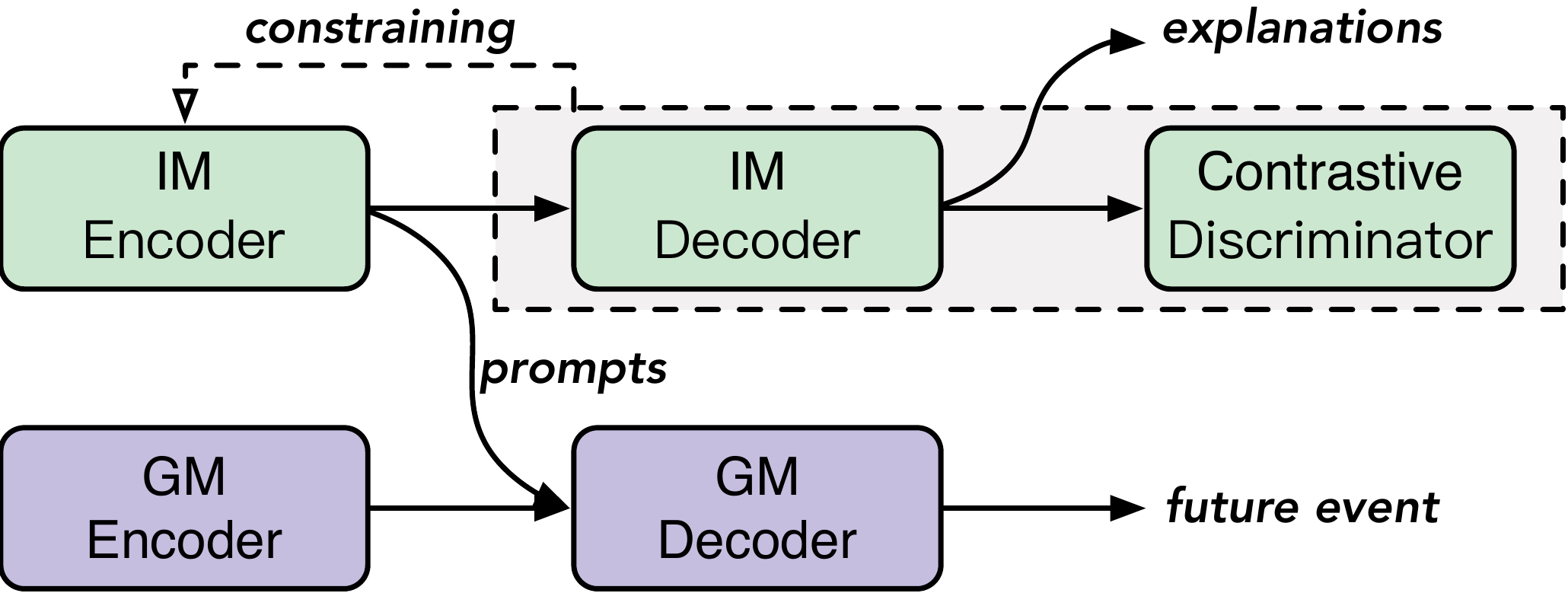}
 	\caption{Future event generation diagram using prompting strategy, where the parameters of \textsc{Im} decoder and discriminator in the dashed block are frozen.}
	\label{fig:FEG}
\end{figure}

Specifically, as aforementioned, given an event and an inferential relation $r_i$, denoted as $x_{\mathcal{I}_i}$, we use \textsc{Im} encoder to get the hidden states from $\text{ENC}_{\mathcal{I}}(x_{\mathcal{I}_i})$ as latent commonsense representation, and the ending positional vectors $\text{ENC}_{\mathcal{I}}(x_{\mathcal{I}_i})_{\langle /s \rangle}$ as prompts to \textsc{Gm}. As there are no gold standard target explanations, we use the pre-trained contrastive discriminator to measure the coherence between input events and decoded explanations. That is to say, we push the latent representations to maintain enough information for the \textsc{Im} decoder to generate an explanation that can get a high score from the contrastive discriminator. The process of future event generation is shown in Figure~\ref{fig:FEG}. But the recurrently decoding on the commonsense representation would bring the non-differentiable issue because of the $argmax$ selecting function, thus we use the straight-through Gumbel Softmax (GS) estimator~\cite{jang2016categorical} to solve it. GS can make it differentiable by providing a continuous relaxation for the one-hot distribution of $argmax$. We can decode the commonsense explanation as follows:
\begin{equation*}
    P(u_t|u_{<t}) = \sigma({\widetilde{\bf H}_{u_t}^{d,l}}{\bf W}+{\bf b})
\end{equation*}
\begin{equation*}
    \widetilde{\bf H}_{u_t}^{d,l} = \text {DEC}_{\mathcal{I}}({\bf H}_{u_{<t}}^{d,l}, \text{ENC}_{\mathcal{I}}(x_{\mathcal{I}_i}))
\end{equation*}
\begin{equation*}
    u_t^p = \text {argmax}(P(u_t|u_{<t}))
\end{equation*}
\begin{equation*}
    {\bf H}_{u_t}^{d,0} = \text {GS}(P(u_t|u_{<t}))\cdot {\bf E}_{V}
\end{equation*}
where the output of $t$-th token in the inference is still selected using $argmax$ but the corresponding states of the token in the input layer of the decoder ${\bf H}_{u_t}^{d,0}$ need to be calculated using GS estimator, and ${\bf E}_V$ is the vocabulary embedding matrix.

To constrain the inferential prompts, we freeze the parameters of the \textsc{Im} decoder and the contrastive discriminator and only update the \textsc{Im} encoder, to minimize the following loss function:
\begin{equation*}
    \mathit{L}_{sc} = -logP({\tilde{\mathbf{I}}_{s}}=0|s=\left \langle x_{\mathcal{I}},u^p\right \rangle)
\end{equation*}
where ${\tilde{\mathbf{I}}_{s}}$ is the estimated label produced by the contrastive discriminator given $x_{\mathcal{I}}$ and commonsense inference $u^p$ generated by the \textsc{Im} decoder. In the end, the overall training loss for the \textsc{Coep} to generate logically coherent future events is defined as follows:
\begin{equation*}
    \mathit{L} = \mathit{L}_{\mathcal{G}} + \mathit{L}_{sc}
\end{equation*}

\section{Experiments}
\subsection{Dataset}
We evaluate our model on a commonsense story dataset~\cite{rashkin-etal-2018-modeling}, which is constructed based on the ROCStories Corpus, containing 14,738 stories that are claimed to have inner psychology of story characters as a chain of mental states to push the story forward.
It has various settings for mental states detection~\cite{tandon2018reasoning, paul2019ranking, otani2019toward}, future event generation \cite{chaturvedi2017story,wang2017integrating}, story telling \cite{yao2019plan, guan2020knowledge} and story cloze test \cite{mostafazadeh-etal-2016-corpus}. Here we create two settings for future event generation and story telling respectively.
As each story consists of 5 sentences of events, for the FEG task, we construct a CommonEvent dataset by unfolding each story and taking the $i$-th sentence as the current event, all previous sentences as history context, and the next sentence as the future event. For storytelling, we simply give the first sentence of each story as a start event and have the models generate all follow-up events.
The statistics of all training datasets are shown in Table \ref{tab: stats-table}, including inferential KG (ATOMIC) that is used to fine-tune the \textsc{Im}, sequential KG (from ConceptNet) which is used to fine-tune the \textsc{Gm}, and CommonEvent that is used to train the whole model \textsc{Coep} to generate future events.

\begin{table}[htbp]
	\centering
	\begin{tabular}{lccc}
		\toprule
		\textbf{} & \#examples & \#words & \#events\_O \\
		\midrule
		\textbf{Inferential KG} & 485,304 & 4.65 & -  \\
		\midrule
		\textbf{Sequential KG} & 39,530 & 2.67 & 24,578 \\
		\midrule
		\textbf{CommonEvent}  & 61,780 & 8.74 & 126 \\
		\bottomrule
	\end{tabular}
	\caption{The statistics of the three datasets. \#examples represents the number of training examples. \#words denotes the average number of words per event. \#events\_O is the number of repeated events in KGs.}
	\label{tab: stats-table}
\end{table}
\subsection{Baselines}
We use the following approaches as baselines as they are commonly used in various generation tasks and have achieved state-of-the-art performance. 
\paragraph{\bf Pointer Generator (Ptr-Gen)} It proposes a hybrid pointer-generator network using coverage to keep track of repeat tokens to discourage repetition based on seq2seq LSTM models~\cite{see2017get}.
\paragraph{\bf GPT-2} It is initialized with a GPT-2 model (small) and trained with an objective to predict the next word. The input to the GPT-2 model is the same as the \textsc{Gm} and the target is the next event sentence in the story on our CommonEvent dataset.
\paragraph{\bf Knowledge-Enhanced GPT-2 (KEG)} It proposes an end-to-end of GPT-2 (small) to incorporate ConceptNet and ATOMIC to implicitly enhance commonsense reasoning comprehension, by reforming the triples in KGs as natural language descriptions to fine-tune the model~\cite{guan2020knowledge}, aiming to generate coherent stories given the opening sentence. It is the most related work with ours since we incorporate the same commonsense KGs with PLMs.
\paragraph{\bf BART} It is initialized from a pre-trained BART (base) model~\cite{lewis-etal-2020-bart} and further fine-tuned on the CommonEvent dataset.
\paragraph{\bf Knowledge-Enhanced BART (KEB)} It is also based on the pre-trained BART (base) model and fine-tuned with reformed ConceptNet and ATOMIC following KEG, the only difference is the language model architecture compared with KEG.

We also introduce several variants of \textsc{Coep} to study the effectiveness of each main component: (1) {\bf \textsc{Coep}-SKG} which omits the sequential KG fine-tuning on \textsc{Gm} to evaluate if implicitly fine-tuning on sequential knowledge improves future event generation. (2) {\bf \textsc{Coep}-PT} which removes prompt training objective $L_{sc}$ to evaluate the effectiveness of the proposed prompt training strategy, which is equivalent to directly concatenating the prompts without any constraint. (3) {\bf \textsc{Coep}-CLS} which omits the classification task $L_{\mathcal{G}}^{cls}$ to verify if the contrastive comprehension can promote event generation.

\subsection{Evaluation Metrics}
We evaluate the experimental results with both automatic metrics and human evaluation. The automatic metrics include:
{\textbf{Perplexity (PPL)}} defined as the exponential average negative log-likelihood evaluating the fluency. Automated metrics to measure the performance of text generation: \textbf{BLEU} \cite{papineni2002bleu}, {\bf ROUGE\_L} \cite{lin2004rouge}, {\bf METEOR} \cite{banerjee2005meteor}, {\bf CIDEr} \cite{vedantam2015cider}, and {\bf BERTScore} \cite{zhang2019BERTScore}\footnote{All these automated metrics are implemented following \cite{hwang2021symbolic}}. 
BLEU is sometimes inappropriate for open-ended text generation because there are multiple plausible future events but only one can be the ground-truth reference in our dataset. 
BLEU scores will become low for large $n$ and it is extremely strict for short text. Since the average number of words in the event descriptions of CommonEvent is about 8.74, which is relatively short compared with summarization or translation tasks, we experimented with $n = 1, 2$.
BERTScore computes the similarity scores for each token in the candidate sentence with each token in the reference sentence using contextual embeddings, which correlates better with human judgments.
\textbf{Repetition-n} \cite{shao2019long} measures the redundancy of stories by computing the average ratio of repetitive $n$-grams in generated stories.
\textbf{Distinct-n} \cite{li2016diversity} measures the generation diversity by the ratio of distinct ones within all generated $n$-grams.

\begin{figure}
	\centering
	\includegraphics[width=\linewidth]{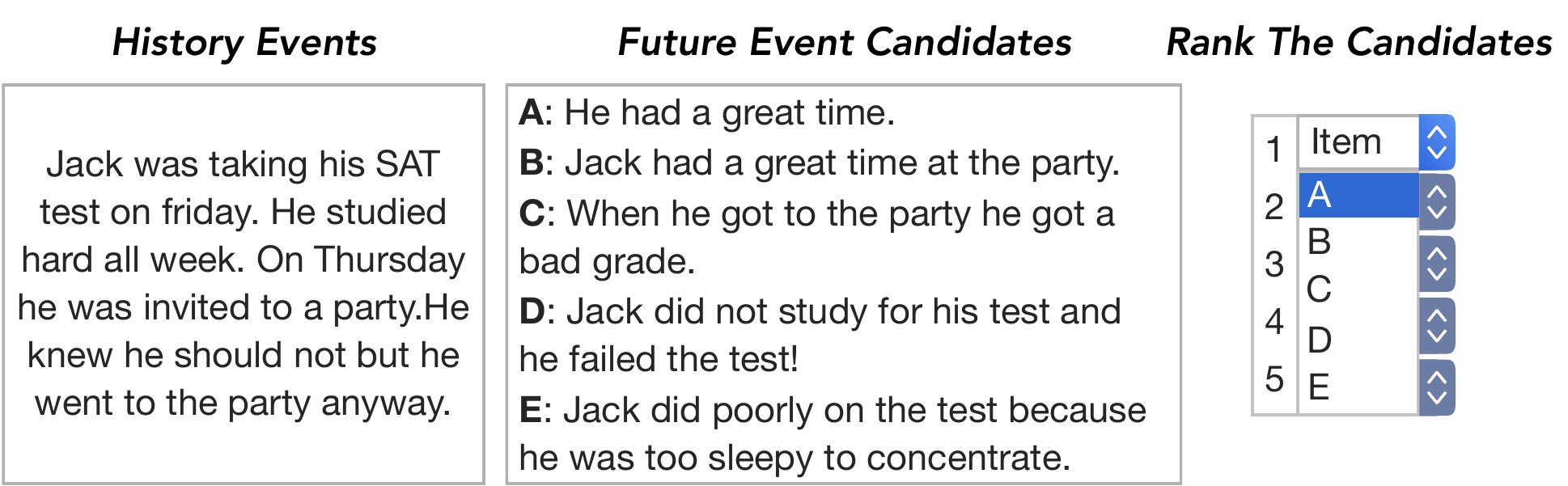}
 	\caption{The illustration of the ranking task for human evaluation of FEG.}
	\label{fig:human-task}
\end{figure}

For human evaluation, we randomly sampled 100 instances from the test set and obtained 400 future events generated by the BART-based models which come top in FEG among the baselines, a variant model w/o PT to investigate the impact of prompt training strategy, and our approach. With the ground-truth, for each instance, we obtain five candidate future events and ask three annotators to rank them based on the logical consistency. The description of the ranking task is shown in Figure~\ref{fig:human-task}.
\textbf{Hit@k} measures the winning rate of each model by computing the percentage of its ranking landing in top $k$ among the candidates. We also use \textbf{Spearman's} $\rho$ \cite{spearman1961general} and the \textbf{Kendall's} $\tau$ \cite{kendall1945treatment} to measure the inter-agreement of annotators.

\subsection{Experimental Setup}
We initialize the encoder-decoder networks with the base version of BART \cite{lewis-etal-2020-bart}. Each is equipped with 6 layers, 768 dimensional hidden states, and 12 attention heads. We inherit the parameters from a pre-trained BART-base model\footnote{We use the pre-trained BART-base model from Hugginface \url{https://huggingface.co/facebook/bart-base}}. The batch size is 32 during fine-tuning on ATOMIC and ConceptNet. While training on event story dataset, we use the batch size of 8 for further extension with 9 commonsense relations. AdamW optimizer with an initial learning rate of 1e-5 and weight decay of 0.01 is used to optimize the models. We use Topk-4 searching strategy to generate future events and commonsense explanations.

\subsection{Evaluation of Future Event Generation}
\begin{table*}
	\centering
	\begin{tabular}{lcccccccc}
		\toprule
		{\textbf{Models}}  & PPL$\downarrow$ & BLEU-1$\uparrow$ & BLEU-2$\uparrow$ & BLEU-4$\uparrow$ & METEOR$\uparrow$  & ROUGE\_L$\uparrow$  & CIDEr$\uparrow$ & BERTScore$\uparrow$ \\
		\midrule
		Ptr-Gen & 25.79 & 5.73 & 0.89 & 0.00  & 4.63 & 6.60  & 0.82 & 38.00 \\
		GPT-2   & 14.51 & 8.35  & 3.98 & 0.67  & 8.95 & 11.45 & 12.29  & 47.61 \\
		BART    & 11.0 & 15.01 & 5.79 & 1.60 & 10.66 & 14.35 & 17.25 & 49.50 \\
		\midrule
		KEG  & 12.17 & 13.41 & 4.37 & 0.80 & 9.75  & 12.57 & 13.82 & 48.63\\
		KEB & 11.38 & 15.38 & 6.13 & 1.75 & 11.01 & 14.52 & 20.25  & 49.91 \\
		\midrule
		\textsc{Coep}   & \textbf{9.62} & \textbf{16.31}  & \textbf{6.74}  & \textbf{1.94} & {\bf 11.95} & \textbf{15.36} & {\bf 25.30}  & {\bf 50.72} \\
		-PT-SKG & 10.80 & 15.62 & 6.29 & 1.79 & 11.27 & 14.88 & 21.19 & 50.17 \\
		-PT      & 10.83 & \underline{15.85} & 6.40 & 1.79 & 11.44 & 14.93 & 21.88 & 50.22 \\
		-SKG      & \underline{10.59} & 15.74 & \underline{6.57} & \underline{1.94} & \underline{11.76} & \underline{15.09} & \underline{24.48} & 50.33 \\
		-CLS      & 11.30 & 15.61 & 6.35 & 1.82 & 11.43 & 14.73 & 24.21 & \underline{50.41} \\
		\bottomrule
	\end{tabular}
	\caption{Automatic evaluation results on FEG task. \textbf{Bold}: the best performance. \underline{Underlined}: the second place.}
	\label{tab:result-FEG}
\end{table*}
\subsubsection{Automatic Evaluation} 
Table \ref{tab:result-FEG} shows the automatic evaluation of FEG performance of all baselines and our approach.
We can see that (1) our model significantly outperforms all the baselines and variants based on all evaluation metrics. (2) BART-based models show obvious superiority compared with both Pointer Generator and GPT-2 models but still suffer from the issue of illogicality, even with conventional KG fine-tuning, which demonstrates the effectiveness of the latent commonsense representations as prompts to future event generation. (3) The highest BERTScore shows that \textsc{Coep} can promote the semantic consistency of generated events, which reveals that our model can effectively capture the commonsense information from KG and apply it to FEG.

Ablation studies on the main components are shown at the bottom of Table \ref{tab:result-FEG}. We can see that (1) without prompt training (-PT) which is equivalent to directly concatenating the commonsense prompts and the preceding events, CIDEr and BERTScore drop rapidly. This verifies the effectiveness of the prompt training strategy to maintain semantic consistency. (2) Fine-tuning \textsc{Gm} on 
sequential KG brings limited improvements.
It is consistent with our claim that implicitly fine-tuning the pre-trained language model with KG lacks effective constraints to control the knowledge inferring on downstream tasks.
(3) The additional classification task in \textsc{Gm} improves the semantic similarity between the events and references, as it uses a related task to enhance the model's contrastive ability.
\subsubsection{Human Evaluation}
\begin{table}[h]
    \centering
    \begin{tabular}{lccc}
        \toprule
        Models & Hit@1 (\%) & Hit@2 (\%) & $\rho$\\
        \midrule
        BART   & 3.34 & 16.70 & 0.23 \\
        KEB & 2.00 & 12.34 & 0.24 \\
        \textsc{Coep}-PT & 2.00 & 33.34 & \textbf{0.29} \\
        \textsc{Coep}        & \textbf{19.33} & \textbf{63.00} & 0.28 \\
        \midrule
        Golden Story & 72.67 & 86.67 & 0.44 \\
        \bottomrule
    \end{tabular}
    \caption{Human evaluation results for FEG.}
    \label{tab:human}
\end{table}

The human evaluation results on generated events are shown in Table \ref{tab:human}, we can see (1) our model achieves a relatively unanimous high rank only second to the ground truth. 19.33 percent of events are rated as the most consistent results, and 63 percent of events are rated as top 2 results. (2) The performance gaps are even larger than that of automatic evaluation. That is, the actual achievements of our proposed model are more than our expectation, the automatic metrics need further improvements. (3) Spearman's $\rho$ calculates the inter agreement between annotators on the rankings of each model and Kendall's $\tau$ computes the agreement on all instances. It seems that the ranking of Golden Story achieves a relatively high consistency among annotators while other models get even performance which is acceptable to consider the human evaluations are convincing. We have an average Kendall's $\tau$ of {\bf 0.412}, which shows moderate agreement among annotators on the sort of 5 candidates in each instance.

\begin{table*}
	\centering
	\begin{tabular}{lccccccc}
		\toprule
		{\textbf{Models}} & BLEU-1$\uparrow$ & BLEU-2$\uparrow$ & METEOR$\uparrow$ & CIDEr$\uparrow$ & BertScore$\uparrow$ & Repetition-4$\downarrow$ & Distinct-4$\uparrow$ \\
		\midrule
		GPT-2 & 17.02 & 5.43 & 11.75 & 6.84 & 50.50 & \underline{5.73} & 90.32 \\
		BART  & \underline{20.53} & 5.86 & \underline{14.23} & 17.01 & 50.32 & 9.44 & 84.01 \\
		KEG      & 17.69 & 5.78 & 12.35 & 8.87 & 50.97 & 6.05 & \underline{91.75} \\
		KEB  & 20.18 & \underline{7.81} & 13.96 & {\bf 17.31} & \underline{51.13} & 8.48 & 81.31 \\
		\midrule
		\textsc{Coep}     & \textbf{22.32} & \textbf{7.85} & \textbf{14.98} & \underline{17.14} & \textbf{52.16} & {\bf 1.96}  & {\bf 98.82} \\
		\bottomrule
	\end{tabular}
	\caption{Automatic evaluation on Story Telling task. \textbf{Bold}: the best performance. \underline{Underlined}: the second place.}
	\label{tab:result-SG}
\end{table*}

\subsection{Evaluation of Story Telling}

To further investigate the commonsense reasoning ability of proposed models, we also provide the performance of several models on storytelling task. Different from GPT-2 based models, which produce the next token autogressively until the end of story, BART-based models generate the next sentence step by step till the last event. Since each story in ROCStories dataset contains 5 sentences, we use the first sentence as the start event and make the models recurrently generate 4 future events to complete it. The results are shown in Table~\ref{tab:result-SG}. Our model achieves the best performance based on almost all metrics except CIDEr, because it relies on low-frequency words rather than the semantic consistency between sentences. The lowest repetition-4 and highest distinct-4 scores indicate that our approach can also generate more diverse and specific events, demonstrating the effectiveness of two sub-model designs combined via prompting.

\subsection{Investigation on Varied Psychology States}
\label{sec:psychology anlysis}
We conduct an additional ablation study on the impact of commonsense prompts based on different inferential relations to investigate the effects of different dimensional psychology states.
We compare the future event generation performance of our approach based on the commonsense prompt from each dimension, as shown in the left columns in Table~\ref{tab: result-ablation-relation}. We can see that among the 9-dimensional commonsense prompts, {\it xEffect} is the most effective one, and even shows better performance than KEB in Table~\ref{tab:result-FEG} which is implicitly enhanced with all dimensions of commonsense knowledge.

\begin{table}[htb]
	\centering
	\begin{tabular}{lcccc}
		\toprule
		\multirow{2}*{Relation} & \multicolumn{2}{c}{Automatic} &  \multicolumn{2}{c}{Human} \\
		\cmidrule(r){2-3}  \cmidrule(r){4-5}
		& BLEU-1/2 & BERTScore & Task\#1 & Task\#2   \\
		\midrule
		xNeed   & 15.35 / 6.12  & \underline{50.12} & 0.55 & 0.22 \\
		xAttr   & 15.23 / 6.06  & 50.09 & 0.62 & 0.48 \\
		xEffect & {\bf 15.61} / {\bf 6.30}  & 50.08 & 0.46 & 0.35 \\
		xReact  & \underline{15.48} / \underline{6.25}  & {\bf 50.15} & 0.47 & 0.39 \\
		xWant   & 15.23 / 6.10  & 49.98 & \underline{0.75} & \underline{0.63} \\
		xIntent & 15.44 / 6.09  & 49.98 & {\bf 0.86} & {\bf 0.68} \\
		\midrule
		oEffect & 15.41 / 6.13 & 50.05 & 0.66 & 0.51\\
		oReact  & 15.21 / 6.10 & 50.09 & 0.57 & 0.49\\
		oWant   & 15.39 / 6.10 & 50.04 & 0.74 & 0.54\\
		\bottomrule
	\end{tabular}
	\caption{Automatic and human evaluations results on FEG task with different commonsense prompts.}
	\label{tab: result-ablation-relation}
\end{table}

\begin{figure*}
	\centering
	\includegraphics[width=0.9\textwidth]{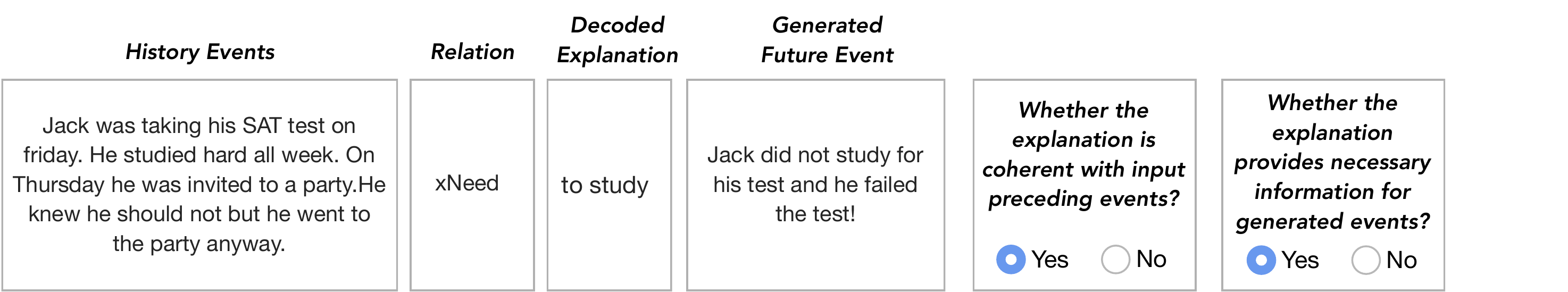}
 	\caption{The illustration of the human evaluation tasks for the correctness of decoded commonsense explanations.}
	\label{fig:human-task2}
\end{figure*}

As the commonsense prompts can also be explained by decoding them into textual commonsense explanations with \textsc{Im} decoder, we further evaluate the commonsense prompts based on the correctness of the textual explanations with human evaluation. We design two tasks for annotators to judge: \textbf{Task \#1}: \textit{whether the explanation is coherent with input preceding events} and \textbf{Task \#2}: \textit{whether the explanation provides necessary information for generated events}, where $1$ stands for yes and $0$ is for no.
Figure~\ref{fig:human-task2} depicts the process of human evaluation given a sample including history events, inferential relation, explanation and the generated future event.
The right columns in Table~\ref{tab: result-ablation-relation} show the average answer scores on randomly sampled 100 instances. We can see that (1) our model can generate reasonable and coherent explanations on 9 dimensions of commonsense relations, especially xIntent, which shows the highest correlation with input events. (2) The explanations serve as a bridge between preceding events and future events, as their score is highly correlated, which well supports our explicitly explainable framework. We find an interesting phenomenon that human evaluations show that the most correlated commonsense explanations come from xIntent relation, but the automatic evaluation results considering only xIntent prompt are rather low. It reveals that although the \textsc{Im} performs well in commonsense reasoning, how to effectively integrate such information in downstream tasks still has a long way to go, which motivates our future work on the model's explainablity.

\section{Case Study}
\subsection{Qualitative Comparison}
\begin{table*}[h]
	\centering
	\begin{tabular}{ll}
		\toprule
		Context: & None.\\
		Current Event: & Ron needed to learn how to \textbf{throw a curveball}.\\
		\midrule
		Future Event: & He ended up consulting his high school's \textbf{coach} for advice.\\
		BART: & He decided to go to the {\itshape{doctor}}.\\
		KEG: & I told my friend I would play with him.\\
		KEB: & He decided to try out for the team.\\
		\textsc{Coep}: & He went to the \textbf{coach} and asked for help.\\
		Explanations: 
		              & xAttr: determined, \underline{curious}; xEffect: \underline{gets exercise};\\
	\midrule
    \specialrule{0em}{1pt}{1pt}
    \midrule
		Context: & Jack was taking his \textbf{SAT test} on friday. He studied hard all week. On Thursday he was invited to a \textbf{party}.\\
		Current Event: & He knew he should not but he \textbf{went to the party} anyway.\\
		\midrule
		Future Event: & Jack \textbf{did poorly} on the test because  he was too sleepy to concentrate.\\
		BART: & He had a {\itshape{great time}}.\\
		KEG: & He had a good weekend and a great time.\\
		KEB: & Jack had a {\itshape{great time}} at the party.\\
		\textsc{Coep}: & Jack did not study for his test and he \textbf{failed} the test! \\
		Explanations: & xNeed: to \underline{study};  xEffect: gets \underline{nervous}\\ 
		\midrule
		\specialrule{0em}{1pt}{1pt}
		\midrule
		Context: & Rachel was doing \textbf{poorly} at math in school.  Her teacher suggested that she seek a tutor to help.\\
		Current Event: & Rachel found a \textbf{tutor} and met with her three times a week.\\
		\midrule
		Future Event: & Rachel's hard work pays off, and she {\bf get the highest mark} on the test.\\
		BART: & The teacher told her that she would not {\itshape{make fun of her}}.\\
		KEG: & She became very good at the math.\\
		KEB: & When she got to school, her teacher told her that she was {\itshape{not good at}} class.\\
		\textsc{Coep}: & After a few months, she was able to \textbf{study very well}.\\
		Explanations: 
		              & xIntent: to get a \underline{better} grade; xAttr: cautious, \underline{concerned};  OEffect: PersonX teacher thinks about PersonX's problem; \\
	    \bottomrule
	\end{tabular}
	\caption{Generated future events from different models. \textbf{Bold} phrases denote \textbf{key} information coherent with inputs. {\itshape{Italic}} words donate {\itshape{improper}} events which are illogical or neutral. \underline{Underlined} words denote \underline{effective} explanations for event generation from \textsc{Coep}.}
	\label{tab:case}
	\vspace{-10pt}
\end{table*}

Table \ref{tab:case} presents several examples with future events generated by various methods, which indicates that our approach consistently generates more reasonable and coherent future events than the baselines. For example, given that \textit{Ron wants to learn about sports (\textbf{curveball})}, \textsc{Coep} will generate a future event suggesting him to \textit{ask a \textbf{coach} for help}.
However, KEG produced a neutral prediction that {\it I told my friend I would play with him.} although it was fine-tuned on the same KGs as ours, which confirms that simply fine-tuning PLMs on KGs will bring forgetting issues on downstream tasks.
We also observe that our approach can also capture the \textbf{turning points}. Considering the second example, the explanation shows that Jack \underline{needs to study} but the event {\it he went to the party the day just before the test} makes he \underline{get nervous} and finally leads to his failure in the test. And an opposite scenario that Rachel {\it did \textbf{poorly} at math} but she {\it found a tutor} and worked hard, our model would give a prediction that {\it after a few month's} hard working, she can {\it study \textbf{well}}. The partial explanations shown in the table also confirm that the \textsc{Im} can provide effective explanations for future event generation.

\subsection{Error Analysis}
\begin{table}
    \centering
    \begin{tabular}{ll}
        \toprule
        Input: & John was really bad boss. \\
        \textsc{Coep}: & He \textbf{did not know what to do} with him.\\
        \midrule
        Input: & Tom always wanted a {\itshape{motorcycle}}. Tom went \\
               & to his local Harley Davidson dealership.\\
        \textsc{Coep}: & Tom picked up a {\itshape{bike}} he liked.\\
        \midrule
        Input: & In 1996, my parents tooks a trip to {\itshape{Europe}}.\\
        \textsc{Coep}: & They went on a trip to {\itshape{Mexico}}.\\
        \midrule
        Input: & Mark was so in love with his girlfriend.\\ 
               & Mark was going to propose to her tonight.\\
               & He took her out to the nicest place in town.\\
               & Mark got down on one knee and ask her \\
               & to marry him.\\
        Next Event: & She said \underline{no} she stopped loving him \\
               & months ago.\\
        \textsc{Coep}: & She said \underline{yes} and Mark was so happy!\\
        \bottomrule
    \end{tabular}
    \caption{Typical errors made by our model. {\bf Bold} words reveal the non-informative phrase. {\itshape{Italic}} words denote the improper synonym replacement or regional inclusion relation. \underline{Underlined} words represent a totally different but reasonable event compared with ground truth.}
    \label{tab:error}
\end{table}

We also present some typical errors made by our model in Table \ref{tab:error}. It shows that although \textsc{Coep} significantly outperforms the baselines and variants in generating reasonable future events, it still makes some errors, such as generic responses {\bf don't know what to do} when given short opening event, improper synonym ({\itshape{bike \& motorcycle}}), chaotic regional relations ({\itshape{Mexico \& Europe}}) and opposite understanding of contexts ({\itshape{yes \& no}} to the same content). Especially the last case, it shows our framework makes yet reasonable but different understanding about preceding events, which is actually not the model's fault, but due to the open ending. It also demonstrates that human evaluation is still necessary for measuring logical coherence in event generation tasks.

\section{Conclusion and Future Work}

In this paper, we propose a novel future event generation framework named \textsc{Coep} which infers commonsense knowledge as soft prompts to enhance the logicality of generated event texts. There are two key components: 1) commonsense Inference Module (\textsc{Im}) and 2) event Generation Module (\textsc{Gm}). We initialize the components by inheriting a BART model pre-trained on a large-scale corpus. Two different KGs are used to fine-tune the models for commonsense reasoning and sequential understanding separately. The soft prompts are supervised by a pre-optimized contrastive discriminator with \textsc{Im} and the corresponding latent representations can be decoded into textual descriptions, which provide explanations and justification for the future event.
Extensive experiments on an open-domain event story dataset show that our model can outperform strong baselines in FEG. Automatic and manual evaluations substantiate the contextual and logical coherence of generated events.

For future work, it would be very interesting to migrate the architecture to a more advanced pretraining model like GPT-3, like achieving the commonsense knowledge in a Few-Shot way or Zero-Shot way to decrease training costs. The pluggable design of the prompting framework is extensible because we can update \textsc{Im} and \textsc{Gm} separately without re-training the whole model, and we would like to explore its application on other generation tasks like summarization and dialogue generation.

\begin{acks}
The work was supported by the National Key Research and Development Program of China (No. 2019YFB1704003), the National Nature Science Foundation of China (No. 62021002 and No. 71690231), Tsinghua BNRist and Beijing Key Laboratory of Industrial Big Data System and Application.
\end{acks}

\newpage

\bibliographystyle{ACM-Reference-Format}
\bibliography{sample-base}




\end{document}